\documentclass[10pt,twocolumn,letterpaper]{article}

\usepackage{cvpr}
\usepackage{times}
\usepackage{epsfig}
\usepackage{graphicx}
\usepackage{amsmath}
\usepackage{amssymb}
\usepackage{mathtools}
\usepackage{booktabs}
\usepackage[bottom]{footmisc}

\usepackage[breaklinks=true,bookmarks=false]{hyperref}

\cvprfinalcopy 

\def\cvprPaperID{****} 

\begin{document}

\title{1st Place Solutions of Waymo Open Dataset Challenge 2020 \\ 2D Object Detection Track}


\newcommand*{\affaddr}[1]{#1} 
\newcommand*{\affmark}[1][*]{\textsuperscript{#1}}
\newcommand*{\email}[1]{\texttt{#1}}
\author{%
	Zehao Huang\affmark[1] \ \ \ \ Zehui Chen\affmark[2*$^\Diamond$] \ \ \ \ Qiaofei Li\affmark[3*$^\Diamond$] \ \ \ \ Hongkai Zhang\affmark[4] \ \ \ \ Naiyan Wang\affmark[1] \\
	\affaddr{\affmark[1]TuSimple}~~~~~
	\affaddr{\affmark[2]Tongji University}~~~~~
	\affaddr{\affmark[3]Xidian University}\\
	\affaddr{\affmark[4]Institute of Computing Technology, Chinese Academy of Sciences
	}\\
	{\small \email{\{zehaohuang18, lovesnowbest, qjliqiaofei, kevin.hkzhang, winsty\}@gmail.com}
}}

\maketitle
\let\thefootnote\relax\footnotetext{* Equal Contribution}
\let\thefootnote\relax\footnotetext{$^\Diamond$ This work was done when Zehui Chen and Qiaofei Li were interns in TuSimple.}
\DeclarePairedDelimiter\ceil{\lceil}{\rceil}
\DeclarePairedDelimiter\floor{\lfloor}{\rfloor}
\newcommand{\thefootnote}{\arabic{footnote}}
\begin{abstract}
In this technical report, we present our solutions of Waymo Open Dataset (WOD) Challenge 2020 - 2D Object Track. We adopt FPN as our basic framework. Cascade RCNN, stacked PAFPN Neck and Double-Head are used for performance improvements. In order to handle the small object detection problem in WOD, we use very large image scales for both training and testing. Using our methods, our team RW-TSDet achieved the 1st place in the 2D Object Detection Track.
\end{abstract}

\section{Datasets}
Waymo Open Dataset (WOD) \cite{sun2019scalability} is a recently public large-scale dataset for autonomous driving research. The dataset provides 1000 scenes for training and validation, and 150 scenes for testing. Each scene contains about 200 frames for each camera and there are 5 high-resolution cameras with resolution 1280$\times$1920 and 886$\times$1920. Overall, the dataset contains about 1.15M images and 9.9M 2D bounding boxes for vehicles, pedestrians and cyclists. Following the official split provided by WOD, we adopt 798 scenes as the training data and 202 scenes for validation in most of cases. All 5 camera images are used for training and evaluation. We do not use other datasets except for ImageNet \cite{deng2009imagenet}.
\section{Methods}
In this section, we will introduce our method with bells and whistles used in this challenge. We will first analyze the object scale distribution of WOD, and then describe our baseline model setting followed by our improvements. All the experiments are conducted in SimpleDet \cite{chen2019simpledet}.
\subsection{Small objects matters in WOD}
\label{smallobject}  
\begin{figure}[t]
	\centering\includegraphics[height=1.4in]{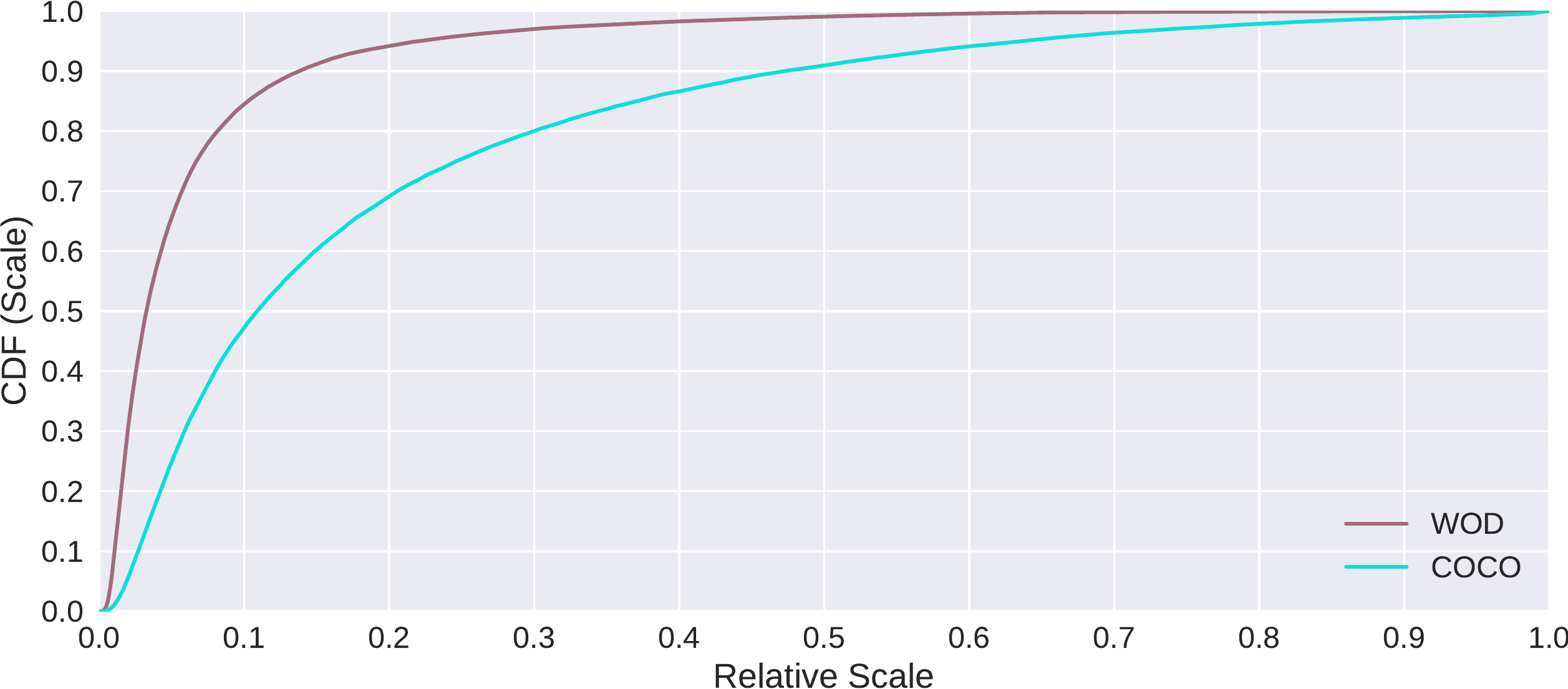}
	\caption{Fraction of bounding boxes in the dataset vs scale of bounding boxes relative to the image.}
	\label{fig:scale}
\end{figure}  
Following SNIP \cite{singh2018analysis}, we compare the scale distribution of WOD with the most common used 2d object detection dataset, COCO \cite{lin2014microsoft}. As show in Fig. \ref{fig:scale}, the relative scale of objects in WOD in about 3 times smaller than COCO.
According to the definition of small objects (area smaller than 32 $\times$ 32) in COCO, about 70\% of objects\footnote{Assume the resolution of images in COCO is about 400 $\times$ 600. We define small object as the bounding box whose relative scale is smaller than $\sqrt{32*32/400/600}=0.065$} in WOD are small objects. Detecting very small objects is a challenging problem in 2d object detection task. Several methods have been proposed to handle this issue, such as image pyramids \cite{singh2018analysis}, feature pyramids \cite{lin2017feature} and so on. In the following, we will describe how we adopt these strategies to improve the detection performances.
\subsection{Baseline model}
We adopt FPN \cite{lin2017feature} with ResNet50-v1b \cite{he2016deep, he2019bag} as the baseline model. According to the observation in Sec. \ref{smallobject}, we modify the default setting of anchor scale and RoI assignment strategy in FPN. The area of anchors are defined as \{$12^2$, $24^2$, $48^2$, $96^2$, $192^2$\} pixels on \{$P_2$, $P_3$, $P_4$, $P_5$, $P_6$\}. As for the RoI assignment strategy in RCNN, we assign an RoI of width $w$ and height $h$ to the level $P_k$ of the feature pyramid by:
\begin{equation}
k = \floor{k_0 + \log_2(\sqrt{wh} / 76)}.
\end{equation}

The input images are resized to a short side of 640 for both training and testing. Random horizontal flip is adopted during training. By default, models are trained in a batch size of 64 on 4 nodes. Each node contains 8 2080Ti GPUs. We adopt SGD with momentum 0.9 and weight decay 1e-4 for training. The learning rate is set to 0.02 for batch size of 16. The linear scaling rule with warmup scheme \cite{goyal2017accurate} is adopted for larger batch size. Cosine decay \cite{loshchilov2016sgdr} is used to attenuate the learning rate over time. We adopt synchronized Batch Normalization (BN) \cite{peng2018megdet} in the backbone, FPN neck, and heads. Note we only calculate BN statistics across GPUs on the same node. We train the models for 8 epochs\footnote{we train about 12 epochs for the baseline model, and 8 epochs for the remain experiments.}. 

For inference, we use the score 0.03 to filter out background bounding boxes and apply the Non-Maximum Suppression (NMS) with the IoU threshold 0.5 per class to get final predictions. The baseline model achieves an mAP of 57.6 at LEVEL 1 and 50.15 at LEVEL 2 among ALL\_NS (the mean over the APs of vehicles, pedestrians and cyclists.)

\subsection{Bells and whistles}
In this section, we show the step-by-step improvements of different components adopted in our methods. Results on WOD validation set can be found in Table. \ref{table:results}. Note that we adopt score first matcher for fast evaluation.

\noindent\textbf{Multi-scale training.} We adopt multi-scale training with random crop. The scale of short side is randomly sampled from \{512, 640, 960, 1280\} and the scale of long edge is fixed as 2000. After rescaling, images are then cropped to a fixed size of 640$\times$960. For image with annotations, we randomly choose an object and jitter the object center as crop center. As for background image, we do random crop.

\noindent\textbf{Large scale testing.} In order to boost the performance of small objects, we adopt a large scale (1280, 1920) for testing. As shown in Table.\ref{table:results}, large scale testing achieves 60.37 mAP at LEVEL 2 among ALL\_NS, which outperforms the small scale results by 6.7.

\noindent\textbf{Cascade RCNN.} WOD adopt different intersection over union (IoU) thresholds to define positives and negatives for different classes, e.g. 0.7 for Vehicles, 0.5 for both Pedestrians and Cyclists. In order to generate high quality detection, we use Cascade RCNN \cite{cai2018cascade}. Following the original implementation, we set the IoU thresholds to 0.5, 0.6 and 0.7 for each RCNN stage respectively. We also try different IoU thresholds, and find that the default setting yields best performance.

\noindent\textbf{Double-Head.} We adopt a Double-Head \cite{wu2019double} method to replace the default fully connected head (\textit{2-fc}) in FPN. For each RCNN stage in Cascade RCNN, a \textit{2-fc} head is used for classification and a convolution head (\textit{conv-head}) with 3 stacked residual blocks is used for bounding boxes regression.

\noindent\textbf{PAFPN.} Following PANet \cite{liu2018path}, we construct PAFPN module by adding bottom-up path aggregation upon FPN. Instead of repeating only once in the original paper, we stack this module 3 times to enhance feature representation. 

\noindent\textbf{Training tricks.} (1) We do not append ground truth into proposals in the RCNN stage during training. (2) Following the training hyper-parameters described in MMDetetcion 2.0 \cite{chen2019mmdetection}, we change the number of proposals after nms from 2000 to 1000.

\noindent\textbf{Multi-scale training with more scales.} We adopt more scales for multi-scale training. The scale of short side is randomly sampled from \{640, 960, 1280, 1600, 1920, 2240, 2560\} and the scale of long edge is fixed as 3840.

\begin{table}[]
	\caption{Step-by-Step improvements on WOD validation set. We report the results of different classes at LEVEL 2.}\label{table:results}
	\vspace{8pt}
	\centering
	\footnotesize
	\begin{tabular}{c|ccc|c}
		\hline
		\multicolumn{1}{c|}{Methods} & Vehicle & Pedestrian & Cyclist & {ALL\_NS} \\\hline
		Baseline       & 50.74 & 64.72 & 34.99& 50.15  \\
		+ Multi-scale training       & 51.67 & 67.19 & 42.16 & 53.67 \\
		+ Large scale testing & 57.33 & 72.90 & 50.88 & 60.37 \\
		+ Cascade RCNN    & 61.10 & 73.67  & 52.00  & 62.26 \\
		+ Doubel Head    & 62.79 & 74.97  & 51.96  & 63.24\\ 
		+ PAFPN & 63.78 & 75.81 & 48.62 & 62.74 \\
		+ Training tricks & 64.21 & 75.56 & 51.51 & 63.76 \\
		+ More scales training & 64.02 & 75.77 & 54.38 & 64.72 \\
		+ Res2Net50-v1b & 64.19 & 76.16 & 54.93 & 65.09 \\
		+ Calibration and flip & 64.98 & 77.40 & 57.98 & 66.79 \\
		+ Scale-aware testing & 66.38 & 78.76 & 61.66 & 68.93 \\
		+ NMS and Voting & 68.99 & 78.72 & 61.20 & 69.64 \\
		+ 3 models ensemble & \textbf{69.81} & \textbf{79.00} & \textbf{62.62} & \textbf{70.48} \\
		\hline
	\end{tabular}
\end{table}

\noindent\textbf{Calibration.} We find that our FPN models tend to predict overconfident classification score on WOD. The scores of many predictions are nearly 1.0. Such score distribution is not suitable for AP metric, since the PR curve may not be smooth at high precision interval in WOD evaluation setting. To optimize for AP, inspired by \cite{guo2017calibration} and \cite{hinton2015distilling}, we calibrate the classification score by adopting a temperature $T=2$ to produce a softer score distribution over classes.

\noindent\textbf{Scale-aware testing.}  Multi-scale testing is a widely adopted trick for 2d object detection. We adopt two scales for testing, (1280, 1920) and (2240, 3360). By analyzing results obtained from different scales, we found that testing large scale image yields higher mAP on small objects\footnote{For vehicles, we define a bounding box whose area smaller than 67 $\times$ 67 as a small object. For pedestrians and cyclists, we set the area threshold to 57 $\times$ 57.} while the small scale one yields higher mAP on large objects. Based on this observation, we degrade the classification score by multiplying a constant value of 0.6 for small scale predictions when testing with (1280, 1920). As for scale (2240, 3360), we decay the score of large scale objects. This simple strategy brings 1.0 improvement than vanilla multi-scale testing.

\noindent\textbf{NMS and Voting.} We set NMS threshold as 0.7,0.5,0.5 for vehicles, pedestrians and cyclists respectively. Bounding box voting \cite{gidaris2015object} is applied after NMS. For simplicity, the threshold of voting is set to the same as NMS threshold. We also try Soft-NMS \cite{bodla2017soft}. Comparing to  valina NMS with bounding box voting, Soft-NMS yields similar results but is inefficient with numerous predictions.

\noindent\textbf{Ensemble.}  We find that the performance between very deep convolutional neural networks, including ResNet101 and ResNet152, and ResNet-50 is not significant. Especially for cyclists, both ResNet101 and ResNet152 yield worse performance than the shallow one. For training efficiency, we adopt three models, Res2Net50-v1b\footnote{\href{https://github.com/Res2Net/Res2Net-PretrainedModels}{https://github.com/Res2Net/Res2Net-PretrainedModels}} \cite{gao2019res2net}, ResNet50-v1d\footnote{\href{https://gluon-cv.mxnet.io/model_zoo/classification.html}{https://gluon-cv.mxnet.io/model\_zoo/classification.html}} \cite{he2019bag} and ResNeXt50\footnote{\href{https://github.com/apache/incubator-mxnet/tree/master/example/image-classification}{https://github.com/apache/incubator-mxnet/tree/master/example/image-classification}} \cite{xie2017aggregated} for ensemble.
We compare the performances of two different ensemble strategies on validation set. The first one is described in PFDet\cite{akiba2018pfdet} and the other one, named Linear-Reweight, is to assign each model a weight of $w_i$ based on its AP rank $k$ on validation set among $n$ candidates:

\begin{equation}
w_i = \theta_0 + (n - k)\cdot\frac{1 - \theta_0}{n-1}.
\end{equation}
Note that $\theta_0$ is set to 0.5 for all classes.
Based on our results, Linear-Reweight is slightly better than the method proposed by PFDet. So we adopt Linear-Reweight as our final ensemble strategy.

\begin{table}[]
	\caption{The performances of expert models and final ensemble model on the WOD validation set. The results are AP at LEVEL 2 for different classes.}\label{table:expert}
	\vspace{8pt}
	\centering
	\footnotesize
	\begin{tabular}{c|ccc|c}
		\hline
		\multicolumn{1}{c|}{Model} & Vehicle & Pedestrian & Cyclist & {ALL\_NS} \\\hline
    	4 expert models ensemble & 68.61 & 78.54 & 62.01 & 69.72 \\ 
		3 models ensemble & 69.81 & 79.00 & 62.62 & 70.48  \\
		Final ensemble & \textbf{70.13} & \textbf{79.07} & \textbf{63.31} & \textbf{70.90} \\ 
		\hline
	\end{tabular}
\end{table}
\subsection{Expert model}
We train several expert models for cyclists since the number of cyclist objects in WOD is only 81k, which is much less than vehicles and pedestrians (9.0M and 2.7M). We do class-aware sampling during training. For fast experiments, we randomly sample 120K images from the whole training dataset each epoch until the ratio between different classes is 1:1:1:0.3 for cyclists, pedestrian, vehicle and background images. We train expert models for 20 epoch. Comparing to models trained on the whole training set, expert models use much less training samples while provide competitive performances. The results can be found in Table.\ref{table:expert}. We combine 4 expert models and 3 models trained on the whole dataset. The final ensemble model yields 70.90 mAP on validation set.

%
%

\section{Final Results}
\begin{table}[]
	\label{table:final}
	\caption{Submission results on WOD testing set. We report the results of different classes at LEVEL 2. The final submission is an ensemble of models trained on the whole dataset and several expert models trained on a subset of the dataset.}\label{table:results}
	\vspace{8pt}
	\centering
	\footnotesize
	\begin{tabular}{c|ccc|c}
		\hline
		\multicolumn{1}{c|}{Methods} & Vehicle & Pedestrian & Cyclist & {ALL\_NS} \\\hline
		Baseline &56.34 &67.63 &36.92 &53.63\\ 
		Final submission &\textbf{76.63} &\textbf{81.72} & \textbf{64.94} & \textbf{74.43} \\
		\hline
	\end{tabular}
\end{table}
For final submission, we train multiple models on the union of training and validation sets, including ResNet50-v1b, ResNet50-v1d, Res2Net50-v1b, HRNetv2p-W32 and HRNetv2p-W48\footnote{\href{https://github.com/HRNet/HRNet-Image-Classification}{https://github.com/HRNet/HRNet-Image-Classification}} \cite{wang2020deep}. All the models are pretrained on ImageNet. Linear-Reweight strategy is adopted for final ensemble. Results are shown in Table.\ref{table:final}. The final model achieves 74.43 at LEVEL 2 among ALL\_NS in WOD testing set.

{\small
\bibliographystyle{ieee_fullname}
\bibliography{egbib}
}

\end{document}